\documentclass{article}

\usepackage[preprint]{neurips_2021}

\usepackage[utf8]{inputenc} 
\usepackage[T1]{fontenc}    
\usepackage{hyperref}       
\usepackage{url}            
\usepackage{booktabs}       
\usepackage{amsfonts}       
\usepackage{nicefrac}       
\usepackage{microtype}      
\usepackage{xcolor}         
\usepackage{amsmath}
\usepackage{array}
\usepackage[export]{adjustbox}

\DeclareMathOperator*{\argmax}{arg\,max}

\title{Towards Continual Entity Learning in Language Models for Conversational Agents}

\author{%
  Ravi Teja Gadde \\
  Amazon\\
  \texttt{gadderav@amazon.com} \\
  \And
  Ivan Bulyko \\
  Amazon \\
 \texttt{ ibbulyko@amazon.com} \\
}

\begin{document}

\maketitle

\begin{abstract}
Neural language models (LM) trained on diverse corpora are known to work well on previously seen entities, however, updating these models with dynamically changing entities such as place names, song titles and shopping items requires re-training from scratch and collecting full sentences containing these entities. We aim to address this issue, by introducing entity-aware language models (EALM), where we integrate entity models trained on catalogues of entities into the pre-trained LMs. Our combined language model adaptively adds information from the entity models into the pre-trained LM depending on the sentence context. Our entity models can be updated independently of the pre-trained LM, enabling us to influence the distribution of entities output by the final LM, without any further training of the pre-trained LM. We show significant perplexity improvements on task-oriented dialogue datasets, especially on long-tailed utterances, with an ability to continually adapt to new entities (to an extent).
\end{abstract}

\section{Introduction}
\label{introduction}

Neural Language Models have become the de facto standard for modeling natural language; however, they still lack some of the flexibilities offered by their n-gram counterparts \citep{brown1992class} in being able to continually update to new entities at a minimal cost \citep{Pusateri2019} or quickly adapt to personalized content \citep{43819}. This issue is more prominent in task-oriented dialogue systems where utterances are typically rich in entities such as place names, shopping items, song titles, person names etc. Thus task-oriented dialogue systems still rely on finite state transducers \citep{mohri2002weighted} and gazetteers \citep{mikheev1999named} in conjunction with neural models for Automatic Speech Recognition (ASR) \citep{le2021contextualized, Zhao2019, 9413962} and Natural Language Understanding (NLU) \citep{kumar2017just, wu-etal-2018-evaluating} tasks. However, these approaches are difficult to scale because of their poor generalization capabilities, high memory footprint and the need for annotated datasets. Overcoming these challenges is of vital importance to task-oriented dialogue systems to shift from complex hybrid approaches currently in use to all neural models optimized using a single objective.

To address some of these issues, we introduce entity-aware language models (EALM), where we attempt to decouple entity models from the pre-trained LMs. We achieve this by linearly interpolating the final layer output representations of the pre-trained LM and independently trained entity models. The interpolation probabilities at each timestep are determined by a contextual fusion layer based on the sentence context. This approach allows us to better predict entities present in text, using the information provided by the entity models along with the distribution learned by the pre-trained LM. Our final language model consists of three components (1) A pre-trained language model, (2) Entity models, and (3) A Contextual fusion layer to combine the pre-trained LM with entity models.

Our entity models are neural LMs trained on catalogues containing entities. They share the input and output layers with the pre-trained LM and these parameters are frozen throughout the training, essentially mapping entities to the pre-trained LM's embedding space. This modular approach allows us to update our entity models with new information, independently of the pre-trained LM. The improved entity models can be directly plugged back into the final LM without any further training. This enables our LM to continually learn new entities without requiring to update the pre-trained LM.

It is important to note that our final LM should not always rely on entity models while predicting entities in text. Entities in natural language can be classified into two categories based on the context in which they appear: (1) fact-based entities that depend on real-world facts, and (2) preference-based entities that depend on external factors such as time, location, speaker, etc. For example, in the utterances \emph{play hello by adele} and \emph{play any song by adele}, artist \emph{adele} in the first utterance can be classified as a fact-based entity and is determined by the utterance context alone (plus some real-world knowledge), whereas, \emph{adele} in the second is a preference-based entity that depends on auxiliary information outside of the utterance. Thus, given the utterance context \emph{play any song by}, the distribution of the next word depends on many external factors. We expect our entity models to reflect this distribution and our LM to learn to use it from the entity models. The LM should ignore the distribution from entity models in all other cases. Unlike \citep{Khandelwal, peters-etal-2019-knowledge, zhang-etal-2019-ernie}, that are desgined to improve the representation of fact-based entities, in this paper, we only focus on preference-based entities.

Our main contributions are as follows: (1) We propose an approach to integrate entity models trained on catalogues of entities into a pre-trained LM, (2) We show that this improves our LM's capability in predicting long-tailed entities (3) We further demonstrate that updating our entity models with new information outperforms the pre-trained LM on these new entities, however, still lags behind the pre-trained LM, if we are able re-train it with full sentences containing these new entities.

\subsection{Our Approach}
\label{approach}

Given utterance context $w_{0...t-1} = (w_0, ..., w_{t-1})$, the goal of an autoregressive language model is to estimate the probability distribution $p(w_t|w_{0...t-1})$ of the next token $w_t$, where $w_0$ is typically the start token $\verb|<s>|$ used to predict the first token. We can factorize this probability distribution as,

\begin{equation}
	\label{eq:lm-factorization}
	p(w_t|w_{0...t-1}) = \sum_{i=0}^{N} p(w_t|w_{0...t-1}, E_i) \cdot p(E_i | w_{0...t-1})
\end{equation}

where $E_0$ represents the pre-trained LM and $E_1, ... E_N$ represent the entity LMs. However, entity models cannot accurately predict $p(w_t|w_{0...t-1}, E_i)$ as they are only trained on entity catalogues. As a result, we use the utterance context $w_{0...t-1}$ to determine the entity context $w_{t-l^*...t-1}$\footnote{$t-l^*$ denotes the start index of the entity in the utterance, if the current phrase is an entity, $t-l^*=0$ otherwise.} containing only entity terms, for every entity $E_i$, at every timestep $t$ and approximate $p(w_t|w_{0...t-1}, E_i)$ as follows,

\begin{equation}
	\label{eq:entity-approx}
	\begin{gathered}
		p(w_t|w_{0...t-1}, E_i) \approx p(w_t | [w_0; w_{t-l^*...t-1}], E_i) \;\;\; \forall i = 1...N \\
		l^* = \argmax_{l \in [0, t-1]} \; p(l|w_{0...t-1}, E_i) \\
	\end{gathered}
\end{equation}

where $[w_0; w_{t-l^*...t-1}] = (w_0, w_{t-l^*}, ..., w_{t-1})$. We assume $[w_0; w_{t-0...t-1}] = w_0$ for notational simplicity. We further simplify Eq.~\ref{eq:entity-approx} using markov assumption on the entity models by only considering $k$ tokens preceding the current timestep, $l^* = \arg \max_{l \in [0, k]} p(l|w_{0...t-1}, E_i) $. We show the utterance contexts of our entity models at different timestamps using an example in Table~\ref{table:entitycontexts}

To convert this into a continuous optimization problem, we rewrite Eq.~\ref{eq:entity-approx} as an expectation over all possible entity contexts and our updated equation looks like,

\begin{equation}
	\label{eq:entity-models}
	p(w_t|w_{0...t-1}, E_i) \approx \sum_{l=0}^{k} p(l|w_{0...t-1}, E_i) \cdot p(w_t | [w_0; w_{t-l...t-1}], E_i) \;\;\; \forall i = 1...N
\end{equation}

\section{Entity-Aware Language Models}
\label{model}

In this section, we describe our proposed Entity-Aware Language Models (EALM). Our LM has three components that are trained sequentially: (1) A pre-trained language model, (2) Entity models and, (3) A contextual fusion layer.  Our LM components and their interactions are outlined in Figure~\ref{fig:ealm}

Given an utterance context  $w_{0...t-1} = (w_0, ..., w_{t-1})$, our LM first converts these tokens into a dense representation  $X = (x_0, ..., x_{t-1})$, where $X \in \mathbb{R}^{t \times d}$, using an input embedding layer. The pre-trained LM and the entity models process this input representation as described below. 

\subsection{Pre-trained Language Model}

We use a multi-layer Transformer decoder similar to \cite{NIPS2017_3f5ee243, radford2018improving} as our pre-trained LM. Given the input representation $X$, the pre-trained LM first adds positional information to it, using a trainable positional embedding matrix $P_e \in \mathbb{R}^{t \times d}$. Our modified input $H_0^P = X + P_e$ is then transformed using a stack of global self-attention and fully connected layers,

\begin{equation}
	\label{eq:pretrained-lm}
		H^P = GlobalAttentionTransformer(H_0^P)
\end{equation}

The last token representation $h_{t-1}^P$ of the final layer's output $H^P$, where $h_{t-1}^P \in \mathbb{R}^d$ and $H^P \in \mathbb{R}^{t \times d}$, is used as an encoding of the utterance context $w_{0...t-1}$. During the pre-training state, $h_{t-1}^P$ is passed through an additional output layer to obtain the probability distribution $p(w_t) = Softmax(h_{t-1}^PW_o)$ of the next token $w_t$, where $W_o \in \mathbb{R}^{d \times |V|}$ is the output embedding matrix shared by all the models.

\subsection{Entity Models}

Unlike the pre-trained LM that takes the entire utterance context  $w_{0...t-1}$  as input, entity models only process the last $l$ tokens, where $l \leq k$, at every timestep, as described in Eq.~\ref{eq:entity-models} and Table~\ref{table:entitycontexts}. We, therefore, use local attention \cite{pmlr-v80-parmar18a} and relative positional embeddings\footnote{We do not have absolute position information while training out entity models on entity catalogues.} \cite{shaw-etal-2018-self} for our entity model self-attention layers. In addition to this, our entity models generate $k+1$ outputs at every timestep, by varying the utterance context length from $0$ through $k$ based on Eq.~\ref{eq:entity-models}. To do this efficiently on modern hardware, we generate the $k+1$ outputs at the same time, sharing computation whenever possible.

To this end, we duplicate the input representation $X^{'} = (x_0, x_{t-k}, ..., x_{t-1})$ of our entity models, $k+1$ times, and obtain a new input representation $H_0^{E_i} \in \mathbb{R}^{k+1 \times k+1 \times d}$, for every entity model $E_i$. Given $H_0^{E_i}$, our entity models generate the last token representations of the final layer, $h_{t-1}^{E_i} \in \mathbb{R}^{k+1 \times d}$ \footnote{Note that $k+1$ here does not represent the time dimension, but rather represent the $k+1$ different outputs obtained by changing the utterance context length from $0$ to $k$.}, by effectively masking the utterance context based on the context length $l$, where $l \in [0, k]$,

\begin{equation}
	\label{eq:entity-lm}
	\begin{gathered}
		h_{t-1}^{E_i} = LocalAttentionTransformer(H_0^{E_i})
	\end{gathered}
\end{equation}

While training the parameters of our entity models on entity catalogues, we only generate one output (instead of $k+1$) and the final layer representation $htrain_{t-1}^{E_i} \in \mathbb{R}^{d}$ is passed through an additional output layer with output embeddings $W_o$. However, unlike the pre-trained LM we freeze the input and output embedding layers, after initializing them with those of the pre-trained LM.

\subsection{Contextual Fusion Layer}
\label{section:contextualfusion}

\begin{figure}
	\centering
	\includegraphics[height=0.4\textheight,width=1.0\linewidth]{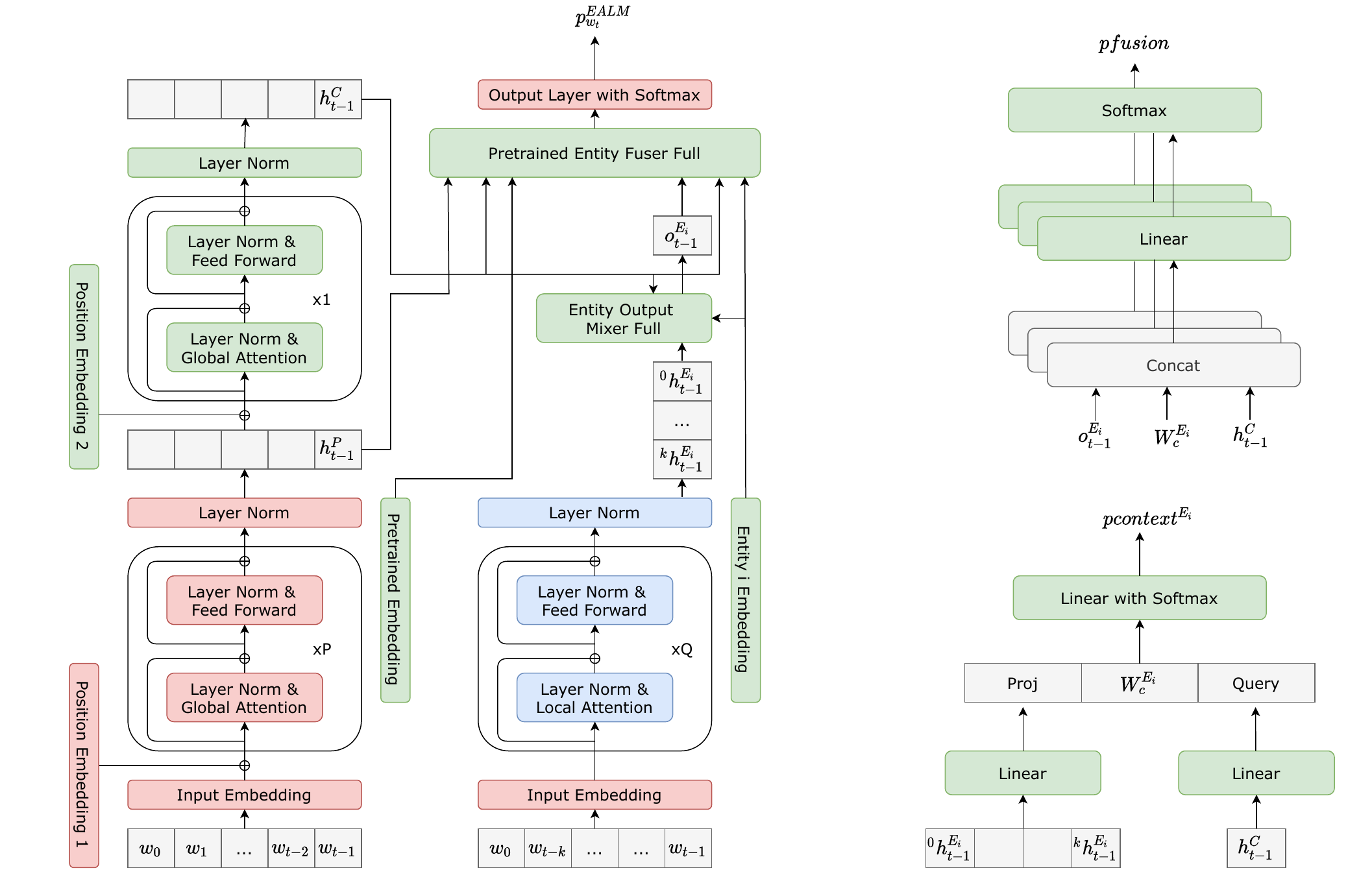}
	\caption{Figure showing the EALM architecture in detail. EALM components and their interactions are shown in the left.  Pre-trained and entity model fuser is shown in the top right and entity output mixer is shown in the bottom right part of the figure. Pre-trained LM parameters are highlighted in red, entity model parameters are in blue and contextual layer parameters are in green. We have only shown one entity model in the figure, additional models follow the same structure.}
	\label{fig:ealm}
\end{figure}

The contextual fusion layer integrates information from the entity models into the pre-trained LM by linearly interpolating their final layer output representations based on the utterance context \footnote{This is slightly different from Eq.~\ref{eq:lm-factorization} where the interpolation is done on the probability distributions of the next token.}. The fusion layer represents the entity models and the pre-trained LM using a class embedding matrix $W_c \in \mathbb{R}^{N+1 \times d}$, where $N$ is the number of entity models. It also creates a new encoding $h_{t-1}^C \in \mathbb{R}^d$ of the utterance context $w_{0...t-1}$ by passing the final layer output $H^P$ of the pre-trained LM through an additional transformer decoder similar to Eq.~\ref{eq:pretrained-lm}. 

The goal of the contextual fusion layer is twofold: (1) Approximate the probability distributions of the next token given by our entity models by identifying their respective entity contexts (Eq.~\ref{eq:entity-models}) and (2)  Determine the probability distribution of the next token given by our final LM by estimating the interpolation weights (Eq.~\ref{eq:lm-factorization}). As mentioned, we use the final layer outputs of our models as a proxy to represent the probability distributions of the next token.

\paragraph{Entity Contexts} Given the output representation $h_{t-1}^{E_i}$  of the entity model $E_i$, its class embedding $W_c^{E_i}$ and the encoding $h_{t-1}^C$ of the utterance context, we first compute the probability distribution $pcontext^{E_i} \in \mathbb{R}^{k+1}$ over all possible entity contexts at the current timestep as shown in Figure~\ref{fig:ealm}. We then create a compact representation  $o_{t-1}^{E_i} \in \mathbb{R}^d$ of the entity model's output as follows,

\begin{equation}
	\label{eq:entity-outputs}
	\begin{gathered}
	pcontext^{E_i} = EntityOutputMixer(W_c^{E_i}, h_{t-1}^{E_i}, h_{t-1}^C) \\
	o_{t-1}^{E_i} = \sum_{l=0}^{k} {^l}pcontext^{E_i} \cdot {^l}h_{t-1}^{E_i} \;\;\; \forall i = 1...N
	\end{gathered}
\end{equation}

where the left superscript $l$ indicates the $l^{th}$ row of the matrix or vector.

\paragraph{Entity Fusion} Given the output representation of the pre-trained LM $h_{t-1}^P$, compact output representations of the entity models  $o_{t-1}^{E_i}$, class embedding matrix $W_c$ and the encoding $h_{t-1}^C$ of the utterance context, we compute the interpolation probabilities $pfusion \in \mathbb{R}^{N+1}$ as shown in Figure~\ref{fig:ealm}. We combine the output representations of the entity models and the pre-trained LM as follows,

\begin{equation}
	\label{eq:entity-fusion}
	\begin{gathered}
	{^i}pfusion = {^i}Softmax(PretrainedEntityFuser(o_{t-1}^{E_i}, W_c^{E_i}, h_{t-1}^C)) \\
	h_{t-1}^{EALM} = \sum_{i=0}^{N} {^i}pfusion \cdot o_{t-1}^{E_i}
	\end{gathered}
\end{equation}

where  $E_0$ represents the pre-trained LM and $o_{t-1}^{E_0} = h_{t-1}^P$.

We pass the interpolated representation $h_{t-1}^{EALM}$ through the output layer with output embeddings $W_o$ to obtain the probability distribution of the next token given by our model.

\begin{equation}
	\label{eq:ealm-probability}
	p^{EALM}(w_t|w_{0...t-1} ) = Softmax(h_{t-1}^{EALM}W_o)
\end{equation}

We freeze the parameters of the pre-trained LM, entity models, input and output embedding layers while training the contextual fusion layer. All our LMs and the Fusion Layer are trained to minimize the negative log-likelihood of the utterances or entities.

\section{Experiments}
\label{experiments}

\subsection{Experimental Setup}

Our training pipeline consists of three stages: (1) Pre-training a language model (2) Training entity models and (3) Training a contextual fusion layer to combine the pre-trained model with entity models. We only use transformer decoder blocks \citep{NIPS2017_3f5ee243} and linear layers in all our networks. We use $T_{layers}$ to represent the number of layers in the transformer, $d_{model}$ for its hidden, input and output dimensions, $d_{ff}$ for the feedforward dimension, $T_{heads}$ as the number of attention heads and $T_{dropout}$ for dropout between the layers. All our models use the AdamW optimizer \citep{conf/iclr/LoshchilovH19} with a weight decay of $0.1$ and a step-wise decaying learning rate scheduler \citep{NEURIPS2019_2f4059ce} with warmup. We use $lr_{warmup}$ to define number of tokens until warmup and $lr_{decay}$ to define the number of tokens after which we decay the learning rate by a factor of $0.9$. We use $lr_{start}$, $lr_{max}$ and $lr_{end}$ to define the starting, maximum and final learning rates respectively. We conduct all our experiments on utterances belonging to Entertainment, Shopping, LocalSearch and Communication domains.

\paragraph{Pre-trained Language Model Training} Our training data consists of transcribed utterances from user interactions with a voice assistant. The total number of tokens available for training is 40 million. All the utterances are de-identified to preserve the privacy of the users. Our pre-trained model parameters are $T_{layers} = 12$,  $d_{model} = 512$, $d_{ff} = 1024$, $T_{heads} = 8$ and $T_{dropout} = 0.1$. The parameters of our learning rate scheduler are $lr_{start} = 1e^{-6}$, $lr_{max} = 6e^{-4}$, $lr_{end} = 1e^{-6}$, $lr_{warmup} = 2^{22}$ and $lr_{decay} = 2^{24}$. Our vocabulary consists of $8192$ character level subword byte-pair encodings trained using \cite{sennrich-etal-2016-neural}. We truncate all utterances that are longer than $32$ subword units after tokenization and use absolute positional embeddings of size $512$ to represent the position of these tokens. We accumulate gradients for two steps before updating the parameters of our network \cite{ott-etal-2018-scaling}. We do not use any bias parameters for the output linear layer before softmax.

\paragraph{Entity Models Training} We used the following entity models for our experiments: (1) song name, (2) album name, (3) celebrity name, (4) video name, (5) shopping item, (6) place name and (7) person name. The size of our entity catalogues ranges from from $2$ million to $10$ million entries depending on the entity model. The entities in each catalogue are assigned a popularity score using some simple heuristics and we sample the entities based on their popularity during training. Our entity model parameters are $T_{layers} = 4\verb|-|6$, $d_{model} = 512$, $d_{ff} = 1024$, $T_{heads} = 4$ and $T_{dropout} = 0.1$. Our learning rate scheduler parameters are same as the pre-trained model except for $lr_{decay} = 2^{25}$. We use a markov dependency length (Eq.~\ref{eq:entity-models}) of $k = 4$ for all our entity models.

\paragraph{Contextual Fusion Layer Training} We only use $8\%$ of our training data to train the contextual fusion layer. Our catalogues cover around $40\%$ of the entities present in this data. We use a one layer transformer with new positional embeddings to represent the utterance context from Section~\ref{section:contextualfusion}. The output dimension of all our linear layers is $512$ except for the ones preceding a softmax layer. We use a dropout of $0.25$ on the entity model outputs and $0.1$ for the rest of the network. The number of epochs used for training is $4$ and our learning rate scheduler parameters are $lr_{start} = 1e^{-6}$, $lr_{max} = 6e^{-5}$, $lr_{end} = 1e^{-7}$, $lr_{warmup} = 2^{22}$ and $lr_{decay} = 2^{22}$. Our batch size is $256$ and gradient accumulation is done for two steps before parameter updates. We freeze all the parameters of the pre-trained LM and entity models during training. Our approach is fully self-supervised once we collect decent sized catalogues, containing some percentage of the entities present in the data used to train this layer. Our method works better if we train this layer on entities not seen by the pre-trained LM, however, we did not do that for this paper. We also found that the contextual fusion layer ignores certain entity models if there is a huge imbalance in the data used to train this layer, especially, for entities appearing in similar contexts such as songs and videos.

\subsection{Results}

We evaluate our model on its ability to represent the long-tail and its effectiveness in continually learning new entities without forgetting old ones. We provide a description of the test sets used for evaluation in Table~\ref{table:testsets}. All our test sets are created from a much bigger general test set unless specified. Our spanish and tailnew test sets are created artificially by replacing the entities in the general test set with $20k$ corresponding entities. We make an assumption that entities are new if they are not present in our training data and catalogues. We evaluated all our test sets on the pre-trained model and Table~\ref{table:pre-trained} shows the relative percentage difference in perplexity of our tail and new test sets compared to the entity seen test set. We use the tail/new test sets as the reference (denominator) for this table.

\begin{table}[ht]
	\caption{Description of our testsets that are de-identified, transcribed and annotated.}
	\label{table:testsets}
	\centering
	\begin{tabular}[t]{l>{\raggedright\arraybackslash}p{0.075\linewidth}>{\raggedright\arraybackslash}p{0.22\linewidth}>{\raggedright\arraybackslash}p{0.2\linewidth}>{\raggedright\arraybackslash}p{0.19\linewidth}}
		\toprule
		Name  & Size & Entites are seen during training & Entities are present in our catalogues & Every utterance contains an entity  \\
		\midrule
		General & $48k$ & some of them & some of them & no \\
		EntitySeen & $18k$  & yes (a few times) & some of them & yes\\
		EntityTail & $8k$  & only individual words & yes & yes\\
		EntityNew & $14k$ & only individual words & no & yes \\
		EntitySpanish & $23k$ & no & no & yes \\
		EntityTailNew  & $23k$ & no & yes & yes \\
		\bottomrule
	\end{tabular}
\end{table}%

\begin{table}[ht]
	\caption{Relative difference in perplexities of our test sets when evaluated on our pre-trained LM.}
	\label{table:pre-trained}
	\centering
	\begin{tabular}[t]{l>{\raggedright\arraybackslash}p{0.1\linewidth}>{\raggedright\arraybackslash}p{0.1\linewidth}>{\raggedright\arraybackslash}p{0.1\linewidth}>{\raggedright\arraybackslash}p{0.13\linewidth}>{\raggedright\arraybackslash}p{0.13\linewidth}}
		\toprule
		General  & EntitySeen & EntityTail & EntityNew & EntitySpanish & EntityTailNew \\
		\midrule
		-11.46\% & 0 \% & 47.55\% & 54.83\% & 85.8\% & 88.75\% \\
		\bottomrule
	\end{tabular}
\end{table}%

\begin{table}[ht]
	\caption{Table showing the effectiveness of EALM on different testsets. All results shown are percentage reduction in perplexity compared to the pretrained LM. We removed the $\%$ sign.} 
	\label{table:ealm}
	\centering
	\begin{tabular}[t]{l>{\raggedright\arraybackslash}p{0.075\linewidth}>{\raggedright\arraybackslash}p{0.07\linewidth}>{\raggedright\arraybackslash}p{0.07\linewidth}>{\raggedright\arraybackslash}p{0.07\linewidth}>{\raggedright\arraybackslash}p{0.07\linewidth}>{\raggedright\arraybackslash}p{0.07\linewidth}>{\raggedright\arraybackslash}p{0.07\linewidth}>{\raggedright\arraybackslash}p{0.07\linewidth}}
		\toprule
		Testset  & All & Song & Album & Celeb & Video & Item & Place & Person  \\
		\midrule
		\midrule
		\multicolumn{9}{c}{Our base EALM. Entity models are trained on full catalogues.}\\ 
		\midrule
		General & 1.56 & 3.77 & 1.88 & 3.32 & 2.81 & 2.72 & 1.91 & 0.99 \\
		EntitySeen & 5.46 & 8.83 & 5.42 & 9.33 & 0.52 & 2.15 & 1.38 & 2.74 \\
		EntityTail & 10.37 & 16.52 & 9.82 & 16.19 & 4.22 & 7.69 & 4.02 & 5.14 \\
		EntityNew & 5.63 & 8.66 & 6.46 & 8.39 & 3.54 & 5.69 & 3.57 & 0.81 \\
		EntitySpanish & 21.9 & 28.32 & 23.24 & 29.32 & 18.92 & 14.13 & 14.24 & 13.22 \\
		\midrule
		\midrule
		\multicolumn{9}{c}{Entity models are re-trained by adding tail and new entities and plugged back into the EALM.}\\
		\midrule
		General & 1.73 & 3.71 & 2.46 & 3.38 & 3.32 & 3.98 & 1.57 & 1.48 \\
		EntitySeen & 5.34 & 8.36 & 5.45 & 8.92 & 0.31 & 2.57 & 1.5 & 2.45 \\
		EntityTail & 14.79 & 19.89 & 13.29 & 20.44 & 6.18 & 13.89 & 7.23 & 5.43  \\
		EntityNew & 14.38 & 16 & 13.66 & 18.03 & 9.35 & 15.73 & 9.54 & 7.2 \\ 
		\midrule
		\midrule
		\multicolumn{9}{c}{Entity models are re-trained by adding spanish entities and plugged back into the EALM.}\\
		\multicolumn{9}{c}{We did not add the tail and new entities for this experiement.}\\
		\midrule
		General & 1.25 & 2.99 & 1.45 & 2.95 & 2.95 & 2.97 & 1.39 & 0.84 \\
		EntitySpanish &  39.12 & 40.53 & 37.93 & 48.14 & 24.37 & 41.21 & 28.03 & 14.02 \\
		\bottomrule
	\end{tabular}
\end{table}%

\subsubsection{Evaluation on Long-Tail}
We first evaluate our EALM trained using our full catalogues on all our test sets. Entities from the test sets are not manually added to the catalogues and the presence of them is purely coincidental. We have divided our test sets into different groups as shown in Table~\ref{table:testsets} based on this information. We show the relative reduction in perplexity obtained by our baseline EALM compared to the pre-trained LM, on all our test sets, in the first part of Table~\ref{table:ealm}. We also show a breakdown of these results by entity type obtained by evaluating only on utterances which contain at least one entity of that type. Note that the tail and new test sets only contain entities where each word is seen at least once in the training data. We do this to avoid evaluating on bad transcriptions with misspelled words.

We can see that our baseline EALM consistently outperforms the standalone pre-trained model on all our test sets. The improvement is much more significant on tail entities ($10.37\%$), that are unseen during training, but are present in our catalogues. Surprisingly, we found that our model also does well on new entities, which are not present in our catalogues. We assume this is because of the regularization effect of our entity models on the pre-trained LM, which probably overfits to seen entities. Our pre-trained model performs very poorly on the spanish entities test set as expected, since these entities contain new words not seen during training combined with the fact that our catalogues contain other spanish entities explains the significant boost seen on this test set.

\subsubsection{Continual Learning of Popular Entities}
Next, we evaluate our model on whether it can continually learn popular entities. To test this, we re-trained all our entity models by adding entities from the tail and new test sets (including dev sets), to the top five percent (ranked by popularity) of our catalogues. We did not retrain the contextual fusion layer. We see a big improvement from $5.63\%$ to $14.38\%$ on utterances containing new entities, refer Table~\ref{table:ealm}. While this is encouraging, we are still significantly worse compared to the performance of our pre-trained LM on the already existing popular entities in the general test set. From Table~\ref{table:pre-trained}, we can infer that the improvement needed on the entity new test set to match the performance of our pre-trained LM on the entity seen test set is close to $54.83\%$. However, retraining the pre-trained LM frequently and collecting utterances containing these entities is expensive.

We also evaluated our method on whether it can learn popular entities from other languages, which is not atypical in task-oriented dialogue systems. We can see that our model achieved a huge overall perplexity reduction of $39.12\%$. Again, this is still significantly lower compared to the $85.8\%$ (refer Table~\ref{table:pre-trained}) improvement needed to match the performance on existing popular entities.

Notice that even though we see a further reduction in perplexity on the general test set when we retrained our entity models with tail and new entities, because of some possible overlap, we see that our improvement on the general test set reduced from $1.56\%$ to $1.25\%$ after adding the spanish entities. This is expected because the contextual fusion layer was trained on previous entity models. We measure this more accurately in the next section. Unlike other current continual learning approaches, that suffer from significant forgetting, our method only showed a minor degradation on the general test set. This can be explained by Table~\ref{table:entityprobs} where we see that the interpolation probabilities of our entity models are almost always capped around $0.05$. We have also observed that this number can only reach $0.2$ to $0.4$ for new/tail entities explaining the difficulty of our system in being able to truly continually learn new entities, and is always limited by the performance of the pre-trained LM on these entities. However, we have seen cases where this number can touch $0.4$ to $0.8$, in a controlled setting, using fewer entity models, a weaker pre-trained LM and some supervision.

\subsubsection{Continual Learning of Tail Entities}
Finally, we evaluate whether we can add entities to the tail of our catalogues and still see significant improvements. To test this, we have created three sets of catalogues containing top $25\%$, $50\%$ and $100\%$ of entities from the original catalogues. To further control our experiments, we removed all entities that are present in the training data and added back $50\%$ of them to these catalogues. We trained two sets of entity models with different random seeds using the $25\%$ catalogue (Entity25, Retrained25) and one set each with the $50\%$ and $100\%$ catalogues (Retrained50, Retrained100). We also artificially created a test set (EntityTailNew) using entities randomly picked from the bottom $75\%$ of our original catalogues. We train our contextual fusion layer using Entity25 and replace it with others for our experiments. We show our results in Table~\ref{table:continual}.

Our results suggest that replacing the entity models comes at a small cost, since the fusion layer is trained on the previous entity models. However, we are still better than the standalone pre-trained model. We are able to continually learn tail entities but the improvements are smaller compared to learning the popular entities. Note that a lot of our tail entities contain both words that are not seen during training and misspelled words, which explains why our pre-trained model does poorly on these entities compared to the EntityNew test set, refer Table~\ref{table:pre-trained}. 

\begin{table}[ht]
	\caption{Table showing percentage reduction in perplexity compared to the pre-trained LM.}
	\label{table:continual}
	\centering
	\begin{tabular}[t]{l>{\raggedright\arraybackslash}p{0.15\linewidth}>{\raggedright\arraybackslash}p{0.15\linewidth}>{\raggedright\arraybackslash}p{0.15\linewidth}>{\raggedright\arraybackslash}p{0.15\linewidth}}
		\toprule
		Test set & EALM trained with Entity25  & EALM + Retrained25  & EALM + Retrained50 & EALM + Retrained100   \\
		\midrule
		General  & 1.55 & 1.32 & 1.33 & 1.33 \\
		EntityTailNew & 14.52 & 14.02 & 16.95 & 17.97 \\ 
		\bottomrule
	\end{tabular}
\end{table}%

\begin{table}[ht]
	\caption{Example showing the relevant utterance context and interpolation probabilities using $k =4$. We have created this utterance artificially and it is not taken from our de-identified customer data.}
	\label{table:entitycontexts}
	\centering
	\begin{tabular}[t]{l>{\raggedright\arraybackslash}p{0.07\linewidth}>{\raggedright\arraybackslash}p{0.07\linewidth}>{\raggedright\arraybackslash}p{0.07\linewidth}>{\raggedright\arraybackslash}p{0.05\linewidth}>{\raggedright\arraybackslash}p{0.07\linewidth}>{\raggedright\arraybackslash}p{0.07\linewidth}>{\raggedright\arraybackslash}p{0.07\linewidth}>{\raggedright\arraybackslash}p{0.07\linewidth}}
		\toprule
		& Play  & a & sky & full & of & stars & by & coldplay  \\
		\midrule
		\midrule
		\multicolumn{9}{c}{Columns indicate the relevant utterance contexts at different timesteps.}\\ 
		\midrule
		Song Model & <s>  & <s> & a & sky & full & of & stars & <s>  \\		
		&   & &  <s> & a & sky & full & of &  \\
		&   & & & <s> & a & sky & full &  \\
		&   & & & & <s> & a & sky  &  \\
		\midrule
		Celeb Model & <s>  & <s> & <s>  & <s>  & <s>  & <s>  & <s>  & <s>   \\	
		\midrule
		Other Models & <s>  & <s> & <s>  & <s>  & <s>  & <s>  & <s>  & <s>   \\		
		\bottomrule
	\end{tabular}
\end{table}%

\begin{table}[ht]
	\caption{Table showing the interpolation probabilities of our EALM based on Eq. \ref{eq:entity-fusion}. 'a sky full of stars' is a song and 'a sky full of stars for a roof' is an album (soundtrack) present in our catalogues.}
	\label{table:entityprobs}
	\centering
	\begin{tabular}[t]{l>{\raggedright\arraybackslash}p{0.07\linewidth}>{\raggedright\arraybackslash}p{0.07\linewidth}>{\raggedright\arraybackslash}p{0.07\linewidth}>{\raggedright\arraybackslash}p{0.07\linewidth}>{\raggedright\arraybackslash}p{0.07\linewidth}>{\raggedright\arraybackslash}p{0.07\linewidth}>{\raggedright\arraybackslash}p{0.07\linewidth}>{\raggedright\arraybackslash}p{0.07\linewidth}}
		\toprule
			& Play  & a & sky & full & of & stars & by & coldplay  \\
		\midrule

		Pre-trained LM & 1 & 0.92 & 0.94 & 0.92 & 0.93 & 0.85 & 0.97 & 0.97   \\	
		Song Model & 0 & 0.02 & 0.03 & 0.03 & 0.03 & 0.06 & 0.01 & 0.01  \\	
		Album Model & 0 & 0.01 & 0.02 & 0.02 & 0.03 & 0.08 & 0.02 & 0.01   \\	
		Celeb Model & 0 & 0.02 & 0.01 & 0.02 & 0 & 0 & 0 & 0.01   \\	
		Video Model & 0 & 0.01 & 0 & 0.01 & 0 & 0.01 & 0 & 0  \\	
		Item Model & 0 & 0.02 & 0 & 0 & 0 & 0 & 0 & 0   \\	
		Place Model & 0 & 0 & 0.01 & 0 & 0 & 0 & 0 & 0  \\	
		Contact Model & 0 & 0.01 & 0 & 0 & 0 & 0 & 0 & 0  \\	

		\bottomrule
	\end{tabular}
\end{table}%
		
\section{Future Work}

Our language models can be directly used for shallow fusion \citep{8639038, cabrera2021language} and n-best hypothesis rescoring \citep{hu2021transformer, shenoy2021adapting} in seq2seq based speech recognition systems. Our approach can also be extended to integrate entity models directly into the decoders of these systems. To improve the recognition of tail entities, we can adapt our entity models based on context such as user, location, season etc \citep{mcgraw2016personalized}. To scale this, we might have to explore sharing parameters across entity models or interpolating tiny models with our entity models. We can also use our approach for quick hotfixes in production systems.

Another interesting line of work would be to use our EALM for intent classification tasks by making use of the interpolation probabilities Eq.~\ref{eq:entity-fusion} as additional cues indicating the presence of entities. We could also consider a harder task of embedding our entity models deep into the pre-trained LM and fine-tune it. Extending our work to bi-directional masked language modeling \citep{devlin-etal-2019-bert} could open up other possibilities such as few-shot entity recognition. Our LM may not have a huge impact on tasks external \citep{NEURIPS2019_4496bf24} to the dialogue systems which benefit more from fact-based entities and their relations.

\section{Related Work}

\paragraph{External Memory} Adding external memory to the neural networks was first proposed in \citep{WestonCB14, NIPS2015_8fb21ee7} where they used attention mechanism \citep{BahdanauCB14} to access memory, based on the current context. \citep{MerityX0S17, GraveJU17} extended this to LMs to increase the probability of recently seen words and later \citep{GraveSecond} scaled this to larger contexts using k-nearest neighbors to retrieve from memory. Recently, \citep{Khandelwal} proposed to learn new information by just populating the memory with more data using a pre-trained LM trained on a small corpus. These methods are orthogonal to our work as they can only improve the prediction of fact-based or recently seen entities and cannot be easily extended to preference-based entities, using only a few nearest neighbors for interpolation or computing attention over millions of entities \citep{8639034}.  

\paragraph{Continual Learning} Continual learning approaches focus on learning new information without forgetting the previously acquired knowledge. Current continual learning approaches can be classified into three categories: (i) regularization \citep{Kirkpatrick3521, chaudhry}, (ii) memory \citep{ChaudhryRRE19, NEURIPS2019_f8d2e80c} and (iii) expansion \citep{rusu2016progressive, Schwarz0LGTPH18} based approaches. They are still an active area of research and require full sentences to learn new entities. It is also not straightforward on how to use these approaches for personalization and contextualization. 

\paragraph{Entity Models} Earlier work on building separate models for entities and combining them with a general language model used annotated data and class n-gram LMs \citep{ward1996class}. \citep{parvez} extended this to neural models, by predicting the entity type along with the next word and improved the representation of tail entities. \citep{liuaugmented} followed a similar approach but instead encoded the type information in a latent variable with the help of type specific vocabularies and did not use any labeled data. However, these approaches only work for entities seen during training and cannot be extended to support personalization or continual learning. Our work resembles \citep{filimonov2020neural}, where they interpolate component models trained on entity catalogues with a pre-trained LM. Unlike our approach, they explicitly reset the hidden states of the entity models, using a binary activation process, to indicate the start of entities in the utterance context and use reinforcement learning to learn the discrete activation policy.

\section{Conclusion}
We proposed a technique to integrate entity models trained on catalogues of entities into pre-trained LMs, significantly improving the perplexity on long-tailed entities. We also showed that our LM can continually learn new entities at a low cost, outperforming the pre-trained LM on these entities, without any signs of forgetting. We discussed the limitations of our continual learning approach in matching the performance of a pre-trained LM directly trained on these new entities. Nevertheless, our LMs can still be used to learn new entities or hotfix trending entities between deployment cycles or in cases where collecting full sentences containing these entities is expensive. Our LMs can be easily extended to support personalization and contextualization in conversational agents.

\section{Broader Impact}

We explore a direction where we can reduce frequent updates to the pre-trained LMs which are expensive to train while also limiting the manual effort needed for collecting the data to train them. Our LMs make efficient use of entity catalogues, compared to the current state-of-the-art approaches that only learn this information through varied data sources. We envision our LMs to have a significant impact in the conversational AI space. Our approach can be used for contextualization and personalization of neural language models in such systems. We can also use it for quick hotfixes of already deployed models. Our method can be extended to remove biases in language models by improving the representation of tailed entities such as artists, items or restaurants. This has to be extensively tested before releasing such models. Correspondingly, our models can also be used to bias certain entities more than others leading to a poorer representation of those entities.

\bibliographystyle{plain}
\bibliography{neurips_2021.bib}

\end{document}